\documentclass[letterpaper, 10 pt, conference]{ieeeconf}  

\IEEEoverridecommandlockouts                              

\usepackage{graphics} 
\usepackage{epsfig} 
\usepackage{mathptmx} 
\usepackage{times} 
\usepackage{amsmath} 
\usepackage{amssymb}  

\usepackage{booktabs}
\usepackage{multirow}
\usepackage{pifont}
\newcommand{\cmark}{\ding{51}}
\usepackage{xcolor}
\usepackage{hyperref}
\usepackage{dsfont}
\usepackage[mathcal]{euscript}

\usepackage{cite}
\usepackage[table]{xcolor}

\title{\LARGE \bf
P$^3$T: Prototypical Point-level Prompt Tuning with Enhanced Generalization for 3D Vision-Language Models
}

\author{
Geunyoung Jung$^{1}$, Soohong Kim$^{1}$, Kyungwoo Song$^{2}$ and Jiyoung Jung$^{1\dagger}$
\thanks{$^{1}$Department of Artificial Intelligence, University of Seoul, South Korea}%
\thanks{$^{2}$Department of Applied Statistics, Yonsei University, South Korea}%
\thanks{$^{\dagger}$Corresponding author}
\thanks{*Email: {\tt\scriptsize $\{$gyjung975, jyjung$\}$@uos.ac.kr}}
}
\begin{document}

\maketitle
\thispagestyle{empty}
\pagestyle{empty}

\begin{abstract}
    With the rise of pre-trained models in the 3D point cloud domain for a wide range of real-world applications, adapting them to downstream tasks has become increasingly important. However, conventional full fine-tuning methods are computationally expensive and storage-intensive. Although prompt tuning has emerged as an efficient alternative, it often suffers from overfitting, thereby compromising generalization capability. To address this issue, we propose Prototypical Point-level Prompt Tuning (P$^3$T), a parameter-efficient prompt tuning method designed for pre-trained 3D vision-language models (VLMs). P$^3$T consists of two components: 1) \textit{Point Prompter}, which generates instance-aware point-level prompts for the input point cloud, and 2) \textit{Text Prompter}, which employs learnable prompts into the input text instead of hand-crafted ones. Since both prompters operate directly on input data, P$^3$T enables task-specific adaptation of 3D VLMs without sacrificing generalizability. Furthermore, to enhance embedding space alignment, which is key to fine-tuning 3D VLMs, we introduce a prototypical loss that reduces intra-category variance. Extensive experiments demonstrate that our method matches or outperforms full fine-tuning in classification and few-shot learning, and further exhibits robust generalization under data shift in the cross-dataset setting. 
    The code is available at \textcolor{violet}{\url{https://github.com/gyjung975/P3T}}.
\end{abstract}

\begin{figure*}[!t]
\centering
    \includegraphics[width=1.\textwidth]{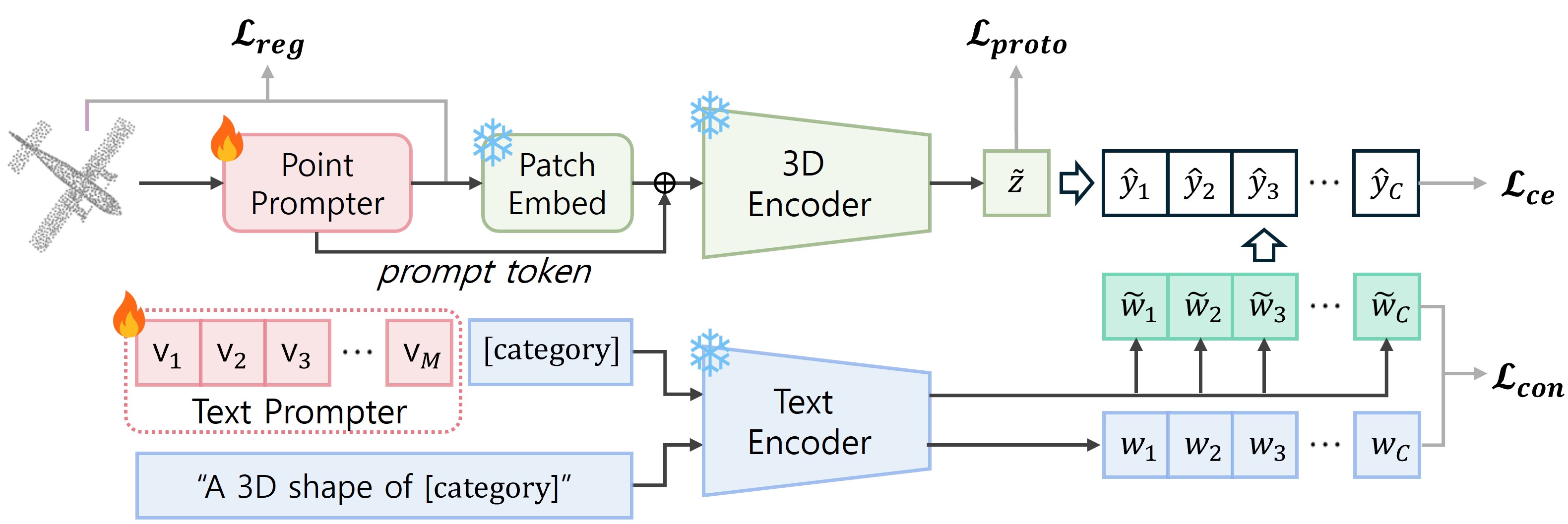}
    \vspace{-7mm}
    \caption{Overview of the P$^3$T framework. The upper part represents the 3D branch with a \textit{Point Prompter}, and the lower part corresponds to the text branch with a \textit{Text Prompter}.}
    \label{fig:framework}
\vspace{-3mm}
\end{figure*}

\section{INTRODUCTION}
\label{sec:intro}
    3D point cloud understanding has become a critical topic in computer vision, drawing considerable attention due to its wide-ranging real-world applications such as autonomous driving and 3D reconstruction. As point clouds represent the most direct and informative form of 3D data, effective processing is essential. However, their inherent irregularity and sparsity pose major challenges. Recent deep learning-based methods~\cite{pointnet, pointcnn, pointconv, point_transformer, pointmixer, pointmlp} have made notable progress by directly operating on point clouds.

    Nowadays, deep learning advanced significantly through large-scale pre-training. In the 3D domain, numerous pre-trained models (PTMs) for point clouds~\cite{pointcontrast, point_mae, point_bert, point_m2ae, pointclustering, pointdif, groupcontrast} also achieved impressive results. 
    Among them, ULIP~\cite{ulip, ulip2} was proposed to learn a shared embedding space across point clouds, images, and text. This approach is inspired by multi-modal model CLIP~\cite{clip}, which exhibits high generalizability and versatility by training on massive image-text pairs. Once pre-trained, fine-tuning yields strong performance on various downstream tasks. However, full fine-tuning is computationally and storage expensive, highlighting the need for parameter-efficient fine-tuning (PEFT) approaches. 

    PEFT involves freezing PTMs and updating only a small number of added learnable parameters. A representative approach is prompt tuning, which inserts learnable parameters, referred to as prompts, into the input of each transformer layer. It shows promising results in both language~\cite{prompt_tuning, prefix_tuning} and image domains~\cite{coop, cocoop, vpt, maple, dapt2, tcp}, often outperforming full fine-tuning while significantly reducing the number of learnable parameters. 
    Recently, several studies have extended prompt tuning to 3D point cloud domain. 
    IDPT~\cite{idpt} and DAPT~\cite{dapt} insert a learnable prompt generation module inside the models. 
    However, both methods are limited to a single modality without any interaction with language, making them incapable of performing zero-shot tasks. 
    PPT~\cite{ppt}, built upon ULIP and capable of zero-shot inference, introduces learnable prompts into the text branch and partially updates layers in the 3D encoder. 
    Since the learnable parameters are either part of or inserted inside the models, all these methods inevitably disrupt the well-aligned embedding space, degrading the inherent generalizability of the PTMs. 

    To address the issue, we propose Prototypical Point-level Prompt Tuning (P$^3$T), a PEFT for pre-trained 3D vision-language models (VLMs). Inspired by VP~\cite{vp} and BlackVIP~\cite{blackvip}, we adopt input-space prompting, i.e., point-level, by directly transforming input point clouds, which means the models remain entirely untouched. P$^3$T consists of two key components: 1) \textit{Point Prompter}, which generates input-dependent point-level prompts from the raw input point cloud, and 2) \textit{Text Prompter}, which introduces learnable prompts into the input text. As illustrated in Fig.~\ref{fig:framework}, all learnable parameters are placed outside the models, thus preserving the generalizability of 3D VLMs. In line with recent strategies that maintain general textual knowledge in hand-crafted prompts~\cite{kgcoop, promptsrc, coprompt}, we also incorporate a consistency loss in the \textit{Text Prompter} to prevent overfitting of the learnable prompts to the target tasks. Furthermore, we observe that the category distinction in the embedding space of the model is unclear. Given that the performance of 3D VLMs heavily depends on the embedding space alignment, we utilize a prototypical loss that reduces intra-category variance. Specifically, we define a prototype for each category as the mean embedding of its train data, and encourage the embedding of each data to be close to its corresponding prototype.

    Extensive experiments across three evaluation settings validate the effectiveness of our method. P$^3$T achieves comparable or even superior performance to full fine-tuning methods on classification and few-shot learning, while significantly reducing the number of learnable parameters. To further assess whether generalizability is preserved after fine-tuning, we conduct a cross-dataset generalization experiment. P$^3$T consistently shows strong generalization performance, highlighting its robustness under data shift.

    The main contributions can be summarized as follows:
    \begin{itemize}[\setlabelwidth{3}]
        \item We propose P$^3$T, a PEFT method that addresses the limitation of existing approaches which degrades generalizability of 3D VLMs. 
        By introducing point-level prompting, P$^3$T avoids disrupting the well-aligned embedding space. In addition, it allows the learnable prompts in the text branch to capture both task-specific and general textual knowledge via a consistency constraint.

        \item To enhance embedding space alignment, we incorporate a prototypical loss that reduces intra-category variance. This particularly benefits real-scanned datasets, where occlusion and noise lead to ambiguous embeddings.

        \item P$^3$T achieves state-of-the-art performance on 3D recognition tasks, surpassing full fine-tuning while requiring far fewer learnable parameters. Cross-dataset evaluation further demonstrates that P$^3$T maintains strong generalizability after fine-tuning.
    \end{itemize}

\section{RELATED WORK}
    \subsection{Multi-modal Pre-training on Point Cloud}
        Recent advances in multi-modal pre-training, especially vision-language models (VLMs)~\cite{clip, align, slip, intervl}, have shown wide applicability. They are typically trained on massive image-text pairs using contrastive learning to learn a shared embedding space between the two modalities. 
        ULIP~\cite{ulip,ulip2} pioneered the extension of VLMs to the 3D point cloud domain, making the first attempt at 3D VLMs. By constructing point cloud-image-text triplets, it enables interaction between point clouds and language, thereby unlocking zero-shot capabilities for various tasks.
        Traditionally, such models are fully fine-tuned for real-world applications after pre-training. 
        However, full fine-tuning is inefficient and may corrupt the rich general knowledge acquired during pre-training. 
        In this paper, we focus on the parameter-efficient adaptation of 3D VLMs.

    \subsection{Prompt Tuning for Pre-trained Models}
        Prompt tuning is an efficient approach for adapting pre-trained models (PTMs) to downstream tasks. It keeps parameters of PTMs frozen and only updates newly added learnable parameters. 
        Originally introduced in the language domain~\cite{power,autoprompt,can,prefix_tuning,bitfit}, it has been studied widely in the image domain~\cite{proda, cocoop, blackvip, pgn, rpo}.
        As the first attempt to apply prompt tuning in the 3D domain, IDPT~\cite{idpt} introduces a lightweight learnable prompt generation module to produce instance-aware prompt, addressing the limitation of conventional input-agnostic static prompts. 
        DAPT~\cite{dapt} further improves the performance by combining adapter tuning~\cite{adapters, adaptformer, unifiedview, fact, lavin}, another PEFT approach. 
        More recently, Point-PEFT~\cite{point_peft} introduces Point-prior Prompt to leverage dataset-specific knowledge. 
        Unlike previous methods, PPT~\cite{ppt} builds upon 3D VLMs and achieves strong performance by optimizing learnable prompts in the text branch and a few layers of the 3D encoder.
        However, adapting the internal parameters of a pre-trained model for specific downstream task often compromises its general knowledge, thereby reducing generalizability.
        In contrast, Point-PRC~\cite{point_prc} applies prompt tuning for domain generalization, at the expense of standard recognition performance.
        Our work is closely related to PPT, but differs in that we explicitly address generalizability, a core strength of 3D VLMs.

\section{METHOD}
\label{sec:method}  
    In this section, we first briefly revisit ULIP~\cite{ulip,ulip2} in Sec.~\ref{sec:pre}. We then introduce P$^3$T consisting of two main components: 1) \textit{Point Prompter} and 2) \textit{Text Prompter} in Sec.~\ref{sec:point_prompter} and~\ref{sec:text_prompter}, respectively. Finally, prototypical loss is described in Sec.~\ref{sec:proto}.
    The overall framework of P$^3$T is illustrated in Fig.~\ref{fig:framework}.

    \begin{figure*}[!t]
    \centering
        \includegraphics[width=1.\textwidth]{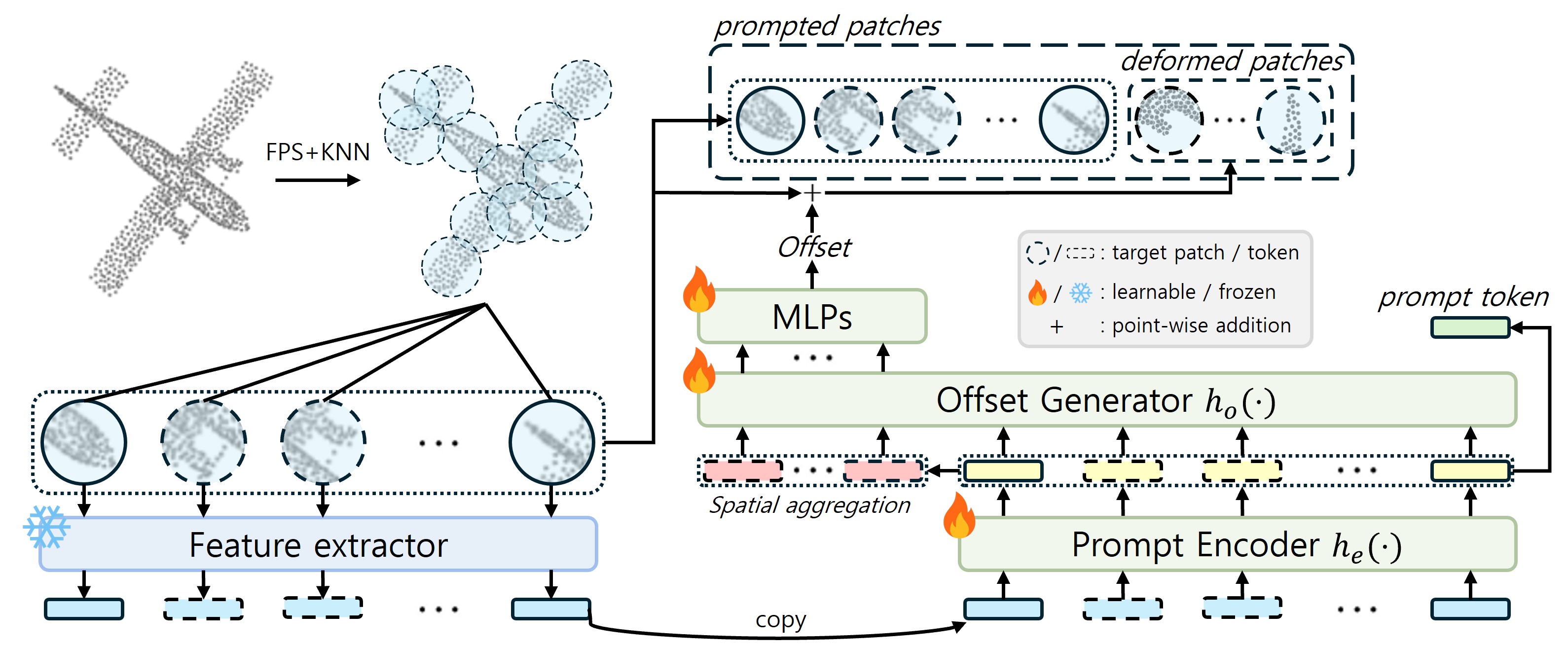}
        \vspace{-7mm}
        \caption{Architecture of \textit{Point Prompter}. It takes $n$ patches of a point cloud, and generates a prompt token and offsets for both local points and center of each patch. The offsets are added to the target patches to create deformed patches, which are then concatenated to original patches.}
    \label{fig:prompter}
    \vspace{-3mm}
    \end{figure*}

    \subsection{Preliminaries}
    \label{sec:pre}
        ULIP is a 3D vision-language model that extends CLIP~\cite{clip} to the point cloud domain. Built on top of CLIP, only 3D encoder $E_P$ is trained via contrastive learning using point cloud-image-text triplets. 
        Given a point cloud $PC$ with $N$ points, it is first divided into $n$ patches $P \in \mathbb{R}^{n \times k \times 3}$ using Farthest Point Sampling (FPS) and K-Nearest Neighbors (KNN), where each patch contains $k$ local points. The following patch embedding layer projects $P$ into patch tokens $X =\{ x_1, x_2, ..., x_n \}$. Then, 3D encoder generates the 3D embedding $z=E_P(X)$. 
        In the text branch, input text for each category, e.g., {\small{``\texttt{a 3D shape of [category]}''}}, is tokenized and projected to text tokens $T_c=\{ t_{\mathrm{SOS}}, t_1, ..., t_{\mathrm{c}}, t_{\mathrm{EOS}} \}$, where $t_c$ denotes the token embedding of category $c$. 
        The text embedding is then obtained via the text encoder as $w_c=E_T(T_c)$.
        Given $C$ categories, the prediction probability is computed as $p(y=c|PC)=\frac{\mathrm{exp}(sim(z, w_c) / \tau)}{\sum^C_{j=1} \mathrm{exp}(sim(z, w_j) / \tau)}$, 
        where $sim(\cdot)$ and $\tau$ are cosine similarity and temperature parameter.

    \subsection{Point Prompter}
    \label{sec:point_prompter}
        Fig.~\ref{fig:prompter} shows the architecture of \textit{Point Prompter}. 
        Without modifying the model, straightforward way to improve performance is to directly manipulate input point clouds. To this end, we apply point-level prompting at the coordinate level. 

        \subsubsection{Target Patch Selection}
        \label{sec:target_selection}
            We select the target patches $P' \subset P$ for prompting based on the idea of ``vulnerable patches''. This is inspired by the ``critical points'' in PointNet~\cite{pointnet}, which refer to key points in a point cloud that remain active after max pooling for global feature.
            In contrast to critical points, our method operates at the patch level and focuses on vulnerable regions that are less informative. 
            It allows vulnerable patches to capture complementary information, thereby enriching the overall representation.
            We first extract patch features $f_P \in \mathbb{R}^{n \times d}$ using feature extractor. For instance-aware prompting, we leverage frozen pre-trained 3D encoder as feature extractor. 
            We define the importance score of each patch as its contribution to the global feature, computed as: 
            \begin{equation}
                s_i = \sum^d_{j=1} \mathds{1}[ i = \arg \max_k(f_P)_{kj}],\quad i=1, ..., n
            \end{equation}
            The bottom $\alpha$ fraction of patches, ranked by importance score, are regarded as vulnerable---that is, less critical for representing the data---and selected as $l$ target patches $P' \in \mathbb{R}^{l \times k \times 3}$, where $l=\mathrm{round}(\alpha \cdot n)$. 
            Note that target patch selection incurs no extra computation, as it relies solely on patch features $f_P$, which are already used in the subsequent stage.

        \subsubsection{Deformed Patches and Prompt Token Generation}
        \label{sec:offset_gen}
            We begin by enhancing patch features to capture multi-scale local geometric structures using the Prompt Encoder $h_e$, which consists of a 3-layer EdgeConv~\cite{dgcnn} followed by a linear layer. Each EdgeConv layer encodes local information at a different scale by dynamically constructing local graphs. The linear layer then projects the resulting multi-scale patch features into offset features $f_o=h_e(f_P)$. 

            Although the offset features of target patches $f'_{o} \subset f_o$ already contain local information, they may lack critical spatial cues at the 3D coordinate level, as EdgeConv defines neighborhoods based on feature similarity. Moreover, since the target patches are selected for their low importance, they tend to be inherently less informative. To address this, we further refine $f'_{o}$ through additional spatial aggregation. 
            Specifically, the refined offset features $\tilde{f}'_{o}$ are computed by max pooling over the offset features of each target patch's neighboring patches.

            These refined offset features, together with $f_o$, are passed to the Offset Generator $h_o$, a 1-layer EdgeConv, to generate the offset tokens $O$.
            A 3-layer MLP is then applied to $O$ to produce patch-wise point offsets:
            \begin{equation}
                [O;\_]=h_o([\tilde{f}'_{o};f_o]),\quad O \in \mathbb{R}^{l\times d}
            \end{equation}
            \begin{equation}
                \delta=\mathrm{Reshape}(\mathrm{MLPs}(O)),\quad \delta \in \mathbb{R}^{l \times k \times 3}
            \end{equation}
            As a result, the deformed patches $\tilde{P}'$ are obtained by patch-wise adding $\delta$ to $P'$, and then concatenated with $P$ to form the prompted patches $\hat{P}$:
            \begin{equation}
                \hat{P} = [P;\tilde{P}'],\quad \text{where }\tilde{P}'=P'+\delta
            \end{equation}
            These are subsequently projected by a patch embedding layer to create the prompted patch tokens $\tilde{X}=\{ \tilde{x}_1, \tilde{x}_2, ..., \tilde{x}_{n+l} \}$. 
            Likewise, a 1-layer MLP generates patch-wise center offsets, which are added to the center points of the target patches to adjust their spatial position.
            
            In addition, following existing prompt tuning methods, we also adopt a prompt token $x_0$ by applying max pooling over $f_o$. 
            Finally, $\tilde{X}$ is fed into the 3D encoder along with a class token and a prompt token for the prompted 3D embedding $\tilde{z}$, computed as:
            \begin{equation}
                \tilde{z}=E_P[x_{cls};x_0;\tilde{X}]
            \end{equation}

        \subsubsection{Point Prompts Regularization}
            Different from fixed-size images, point clouds do not follow any inherent spatial structure or resolution.
            To prevent deformed patches from drifting far from the original point cloud or becoming too large, we constrain their position and size within predefined thresholds.
            Specifically, we define $\mu_Q$ as the centroid of a point set $Q$, and the patch size $\mathcal{D}(Q)$ as the maximum pairwise distance between points in $Q$, i.e., $\mathcal{D}(Q)=\max_{x, y \in Q} \Vert x-y \Vert_2$.
            The threshold for position is set to the maximum distance between the centroids of the original patches and the global centroid: $H=\max_i \Vert \mu_{P_i} - \mu_{PC} \Vert_2$. 
            Similarly, the threshold for patch size is defined as the maximum among all original patches: $G=\max_i \mathcal{D}(P_i)$.
            The resulting regularization loss is formulated as follows:
            \begin{equation}
            \begin{split}
                \mathcal{L}_{reg}&=\frac{1}{l}\sum_{i=1}^l \max(\mathcal{D}(\tilde{P}'_i) - G, 0) \\
                &\quad+ \max(\Vert \mu_{\tilde{P}'_i} - \mu_{PC} \Vert_2 - H, 0),
            \end{split}
            \end{equation}
    
    \subsection{Text Prompter}
    \label{sec:text_prompter}
        Following CoOp~\cite{coop} and PPT~\cite{ppt}, we also apply prompt tuning to the text branch, as illustrated in the lower part of Fig.~\ref{fig:framework}.
        The text tokens of a fixed hand-crafted prompt are replaced with $M$ learnable context vectors $V=\{ v_1, v_2, ..., v_M \}$.
        The text encoder then processes the following tokens $\tilde{T}_c = [V; t_c]$, where $t_c$ denotes the token embedding of category $c$, to generate the prompted text embedding $\tilde{w}_c=E_T(\tilde{T}_c)$. 
        The $M$ learnable tokens are optimized for task-specific text embedding.
        The modified prediction probability with the prompted 3D and text embeddings is computed as:
        \begin{equation}
            p(y=c|PC)=\frac{\mathrm{exp}(sim(\tilde{z}, \tilde{w}_c) / \tau)}{\sum^C_{j=1} \mathrm{exp}(sim(\tilde{z}, \tilde{w}_j) / \tau)}
        \label{eq:ce}
        \end{equation}

        \subsubsection{Text Prompts Regularization}
            Although learnable prompts are effective for adapting models to target tasks, they are prone to overfitting.
            In contrast, hand-crafted prompts contain general knowledge, enabling zero-shot capability across diverse tasks. To balance task-specific adaptation and generalization, we incorporate a consistency loss that minimizes the discrepancy between $w$ and $\tilde{w}$. 
            We compute the cosine similarity as the consistency loss:
            \begin{equation}
                \mathcal{L}_{con} = \frac{1}{C} \sum^C_{i=1} 1 - \frac{w_i \cdot \tilde{w}_i}{\Vert w_i \Vert \ \Vert \tilde{w}_i \Vert}
            \end{equation}

    \subsection{Prototypical Loss}
    \label{sec:proto}
        The primary objective of prompt tuning is to align the embedding spaces of the two modalities, point cloud and text. 
        However, 
        point cloud embeddings are often not well separated across categories. 
        To cluster data from the same category together and make the distinction between different categories clear, we introduce a prototypical loss inspired by PROTONET~\cite{protonet}. 
        This loss encourages each embedding to be close to its corresponding category prototype by minimizing intra-category distances.
        We define the prototype $r_c$ for each category $c$ using the train data. 
        The prototypes are pre-computed before fine-tuning, and kept fixed during training.
        The prototypical loss is formulated as:
        \begin{equation}
            \mathcal{L}_{proto}=1-\frac{\tilde{z} \cdot r_{c\scriptscriptstyle(\tilde{z})}}{\Vert \tilde{z} \Vert \Vert r_{c\scriptscriptstyle(\tilde{z})} \Vert},\quad \text{where } r_c = \frac{1}{\vert S_c \vert}\sum_{(PC_i, y_i) \in S_c} z_i
        \end{equation}
        Here, $S_c$ is the set of train data with category $c$, and $z_i$ denotes the 3D embedding of $PC_i$. $c\scriptstyle(\tilde{z})$ refers to the category of $\tilde{z}$.
        
        \noindent\textbf{Final Loss.}
            The final loss combines three regularization terms with a cross-entropy loss $\mathcal{L}_{ce}$, defined as the negative log-likelihood of the prediction probability in (\ref{eq:ce}).
            \begin{equation}
                \mathcal{L} = \mathcal{L}_{ce} + \beta \cdot \mathcal{L}_{proto} + \gamma \cdot \mathcal{L}_{reg} + \lambda \cdot \mathcal{L}_{con}
            \end{equation}

    \begin{table}[!t]
        \caption{Classification accuracies (\%) on two datasets. \#LP denotes the number of learnable parameters. $^*$ indicates the performance under PPT setting. 
        }
        \label{tab:cls}
        \vspace{-2mm}
        \centering
        \begin{tabular}{l c c c c c}
            \toprule
            \multirow{2}{*}{Method} & \multirow{2}{*}{\#LP (M)} & \multirow{2}{*}{MN40} & \multicolumn{3}{c}{ScanObjectNN} \\
            \cmidrule(lr){4-6} 
            & & & ONLY & BG & PB \\
            \midrule
            \vspace{-3.5mm} \\
            \multicolumn{6}{c}{\emph{Supervised Learning}} \\
            PointNet~\cite{pointnet} & 3.5 & 89.2 & 79.2 & 73.3 & 68.0 \\
            PointNet++~\cite{pointnet++} & 1.5 & 90.7 & 84.3 & 82.3 & 77.9 \\
            PointCNN~\cite{pointcnn} & 0.6 & 92.2 & 85.5 & 86.1 & 78.5 \\
            DGCNN~\cite{dgcnn} & 1.8 & 92.9 & 86.2 & 82.8 & 78.1 \\
            MVTN~\cite{mvtn} & 11.2 & 93.8 & \textbf{92.3} & \textbf{92.6} & 82.8 \\
            PointMLP~\cite{pointmlp} & 12.6 & \textbf{94.1} & -- & -- & \textbf{85.4} \\
            \midrule
            \vspace{-3.5mm} \\
            \multicolumn{6}{c}{\emph{Self-Supervised Pre-training + Full fine-tuning}} \\
            OcCo~\cite{occo} & 22.1 & 92.1 & 85.5 & 84.9 & 78.8 \\
            CrossPoint~\cite{crosspoint} & 27.7 & 90.3 & -- & 81.7 & -- \\
            Point-BERT~\cite{point_bert} & 22.1 & 92.7 & 88.1 & 87.4 & 83.1 \\
            MaskPoint~\cite{maskpoint} & 22.1 & 92.6 & 89.3 & 89.7 & 84.6 \\
            Point-MAE~\cite{point_mae} & 22.1 & 93.2 & 88.3 & 90.0 & 85.2 \\
            Point-M2AE~\cite{point_m2ae} & 15.3 & 93.4 & 88.8 & 91.2 & 86.4\\
            PointGPT~\cite{pointgpt} & 29.2 & 93.3 & 90.0 & 91.6 & 86.9 \\
            ULIP~\cite{ulip} & 22.1 & 92.6 & 89.2 & 91.7 & 84.7 \\
            ULIP-2~\cite{ulip2} & 22.1 & 92.7 & 90.9 & 91.9 & 85.0 \\
            ACT~\cite{act} & 22.1 & 93.6 & 91.9 & 93.3 & 88.2 \\
            \textsc{ReCon}~\cite{recon} & 43.6 & \textbf{93.9} & \textbf{93.1} & \textbf{94.2} & \textbf{89.7} \\
            \midrule
            \vspace{-3.5mm} \\
            \multicolumn{6}{c}{\emph{Self-Supervised Pre-training + Parameter-efficient fine-tuning}} \\
            IDPT~\cite{idpt} & 1.3 & 92.2 & 85.4 & 84.9 & 80.3 \\
            DAPT~\cite{dapt} & 0.8 & 92.4 & 87.1 & 87.1 & 82.3 \\
            Point-PRC~\cite{point_prc} & 0.02 & 90.6 & 87.7 & 89.0 & 79.5 \\
            PPT~\cite{ppt} & 1.8 & 93.1 & 91.8 & 93.7 & 87.2 \\
            P$^3$T (Ours) & 2.0 & \textbf{94.0} & \textbf{93.1} & \textbf{95.2} & \textbf{88.1} \\
            \rowcolor{gray!20}
            Point-PRC$^*$~\cite{point_prc} & 0.02 & 91.1 & 89.3 & 90.2 & 81.9 \\
            \rowcolor{gray!20}
            PPT$^*$~\cite{ppt} &  1.8 & \textbf{94.1} & 93.1 & 95.4 & 89.1 \\
            \rowcolor{gray!20}
            P$^3$T$^*$ (Ours) & 2.0 & \textbf{94.1} & \textbf{93.5} & \textbf{96.4} & \textbf{89.6} \\
            \bottomrule
        \end{tabular}
    \vspace{-3mm}
    \end{table}

\section{EXPERIMENTS}
\label{sec:exp}
        We evaluate P$^3$T on three tasks: classification, few-shot learning, and cross-dataset generalization. The first two tasks assess downstream performance, while the last one evaluates the generalizability of our method.

    \noindent\textbf{Datasets.}
        We use two point cloud datasets, ModelNet40~\cite{modelnet} (MN40) and ScanObjectNN~\cite{scanobjectnn}.
        MN40 is a synthetic dataset with complete and noise-free data, covering 40 categories.
        ScanObjectNN is a real-scanned dataset with 15 categories, and unlike MN40, it has many missing points and noise.
        It is divided into three splits---OBJ\_ONLY (ONLY), OBJ\_BG (BG), and PB---according to the difficulty level. 

        In the cross-dataset generalization, we adopt Objaverse-LVIS (LVIS), which is a subset of the large-scale Objaverse~\cite{objaverse} dataset, for source dataset.
        LVIS contains $\sim$46k data covering $\sim$1.2k categories. Since it does not provide official train and test split, we randomly divide the data into train and test sets using an 8:2 ratio within each category.

    \noindent\textbf{Implementation Details.}
        For all PEFT baselines, we adopt ULIP-2 (Point-BERT)~\cite{ulip2} pre-trained on ShapeNet~\cite{shapenet} as the target 3D vision-language model.
        Its 3D encoder, Point-BERT, is used as the frozen feature extractor of P$^3$T. 
        Only the \textit{Point Prompter}, excluding the feature extractor, and the learnable prompts in the \textit{Text Prompter} are trained. 
        We uniformly sample 1,024 points from each point cloud.

        \begin{figure}[!t]
            \centering
            \includegraphics[width=0.48\textwidth]{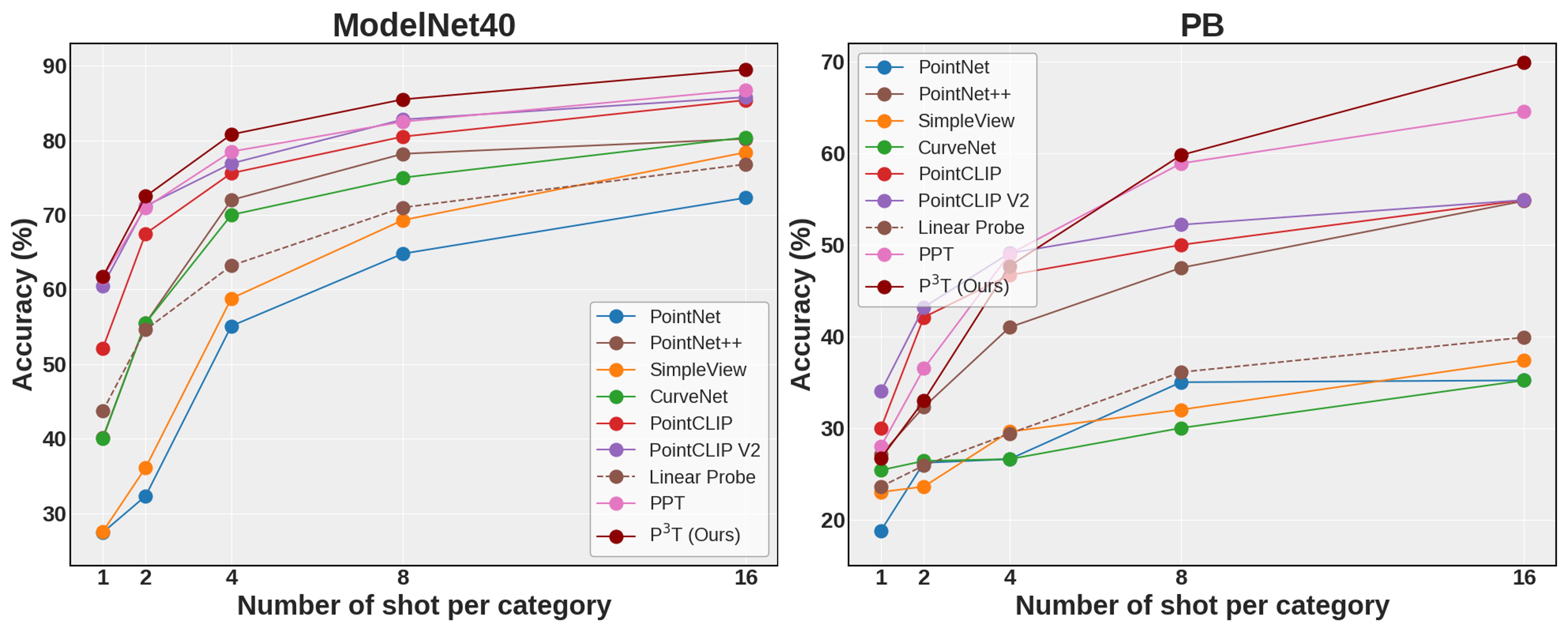}
            \vspace{-7mm}
            \caption{Few-shot classification results on the ModelNet40 and PB datasets. Our method shows strong performance even in the few-shot regime, highlighting its data efficiency.}
        \label{fig:fewshot}
        \vspace{-2mm}
        \end{figure}

        \begin{table}[!t]
            \caption{Results of cross-dataset generalization. P$^3$T achieves the highest average target performance, indicating its outstanding generalizability. ZS denotes zero-shot setting.}
            \label{tab:cross}
            \vspace{-2mm}
            \centering
            \begin{tabular}{lcccccc}
                \toprule
                \multirow{2}{*}{Method} & Source & \multicolumn{5}{c}{Target}       \\
                \cmidrule(lr){2-2} \cmidrule(lr){3-7} 
                 & LVIS   & MN40 & ONLY & BG & PB & Avg. \\
                \midrule
                ULIP-2 (ZS) & 18.1 & 59.8 & 32.2 & 32.0 & 25.3 & 37.3 \\
                \midrule
                ULIP-2    & 38.4 & 58.3 & 23.4 & 26.3 & 16.6 & 31.2 \\
                PPT       & \textbf{41.0} & 61.1 & 39.8 & 40.3 & 26.2 & 41.9 \\
                Point-PRC & 39.3 & \textbf{66.4} & 45.1 & 44.9 & 30.8 & 46.8 \\
                P$^3$T    & 39.6 & 62.4 & \textbf{52.8} & \textbf{48.7} & \textbf{37.0} & \textbf{50.2} \\
                \bottomrule
            \end{tabular}
        \vspace{-3mm}
        \end{table}

    \subsection{3D Object Classification}
        Tab.~\ref{tab:cls} shows classification performance on MN40 and ScanObjectNN datasets. On MN40, P$^3$T achieves 94.0\% accuracy, outperforming all full fine-tuning methods except PointMLP, while using significantly fewer learnable parameters. 
        Compared to ULIP-2, our method performs better with only 2M learnable parameters which is 91\% fewer.
        P$^3$T also exceeds our baseline PPT~\cite{ppt} by 0.9\%, even though PPT updates internal layers of the model.
        On the challenging real-world dataset ScanObjectNN, P$^3$T achieves state-of-the-art accuracy on ONLY (93.1\%) and BG (95.2\%) splits, and competitive performance on PB (88.1\%). 
        The gray area at the bottom, marked with an asterisk ($^*$), shows the results under the default setting in the PPT paper. The same 1,024 points are used for each point cloud, but they are divided into a larger number of patches. This leads to performance improvements, while also increasing computational cost. In this setting, our method also yields substantial performance gains, reaching new state-of-the-art results.

        These results demonstrate effectiveness and efficiency of our method, regardless of the type of dataset.
        Notably, all improvements are obtained without updating any parameters of pre-trained model, relying solely on input-space prompting.

    \subsection{Few-shot Learning}
        The few-shot learning performance on MN40 and PB is illustrated in Fig.~\ref{fig:fewshot}. We randomly sample 1, 2, 4, 8, 16 shots per category for training, and evaluate on the full test set.
        The results are averaged over three runs.
        P$^3$T consistently outperforms existing methods across all shot settings on MN40, demonstrating both parameter and data efficiency. 
        On PB, while P$^3$T underperforms PPT and PointCLIP V2~\cite{pointclipv2} up to 4-shot, it begins to surpass them by a large margin from 8-shot. In particular, it achieves 69.9\% with 16-shot, outperforming PPT by 5.3 points.
        Since PB is the most challenging split with highly noisy and incomplete point clouds, random sampling may yield non-representative samples, resulting in less reliable prototype construction.
        Consequently, under extremely low-shot settings, category prototypes tend to be unstable, which leads to slightly lower performance.

    \begin{figure}[!t]
        \centering
        \includegraphics[width=0.5\linewidth]{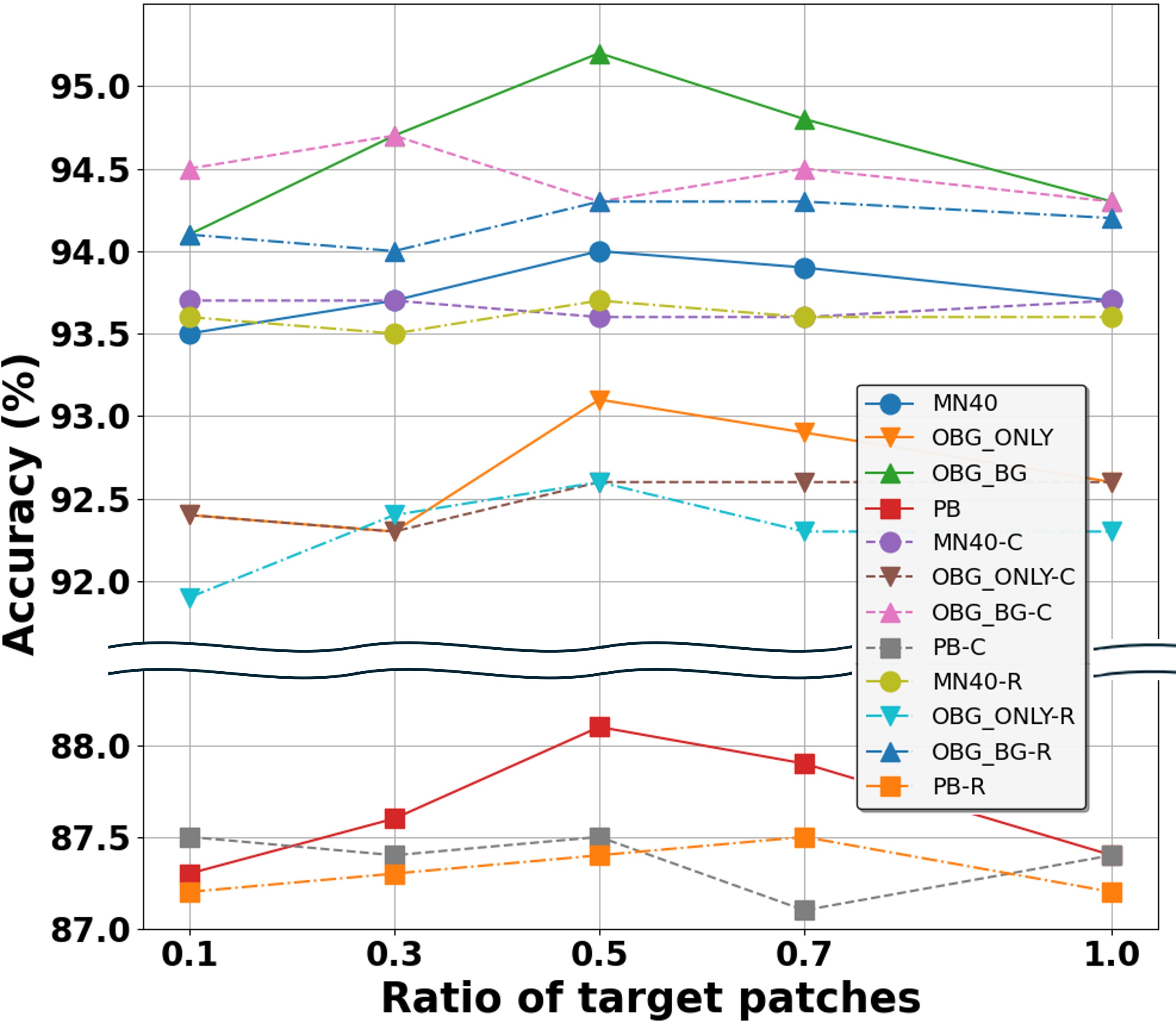}
        \vspace{-3mm}
        \caption{Performance across prompt ratio $\alpha$. The suffix ``-C" and ``-R" indicate the selection of critical and random patches, respectively. Prompting vulnerable patches leads to the best performance at $\alpha=0.5$.}
        \label{fig:ratio}
    \vspace{-3mm}
    \end{figure}

    \subsection{Cross-Dataset Generalization}
        To assess the generalizability of our method under data shift, we conduct a cross-dataset generalization experiment. The models are trained on a source dataset and directly evaluated on unseen target datasets without any fine-tuning. We use LVIS as the source dataset, and MN40 and ScanObjectNN as the target datasets. Due to its large number of data and categories, LVIS serves as a suitable source dataset.

        Tab.~\ref{tab:cross} reports the accuracies on the source dataset and each target dataset.
        While ULIP-2 improves performance on the source dataset at the cost of target performance, P$^3$T improves both simultaneously.
        Compared to PPT, it performs marginally worse on the source dataset, but significantly outperforms PPT on all four target datasets.
        On average, we surpass PPT by 8.3\%, with particularly substantial gains on ScanObjectNN. 
        For Point-PRC~\cite{point_prc}, although it is tailored for domain generalization and performs better on MN40, P$^3$T achieves consistently higher performance on all other datasets, resulting in 3.4\% higher average performance on the target datasets. 
        Note that Point-PRC attains generalizability by sacrificing standard classification performance, as evidenced by Tab.~\ref{tab:cls}. 
        These results suggest that placing learnable modules entirely outside the model facilitates stronger generalization. 
        We further attribute this effect to point-level prompting on the input point cloud, which directly alters the input distribution and helps preserve generalizability under data shift.

    \begin{table}[!t]
        \centering
        \caption{Ablation study on Point and Text Prompters. Each Prompter individually improves performance, and their combination achieves the best results across all datasets.}
        \label{tab:ab_prompters}
        \vspace{-2mm}
        \setlength\tabcolsep{1.5mm}
        \begin{tabular}{ccccccc}
            \toprule
            \textit{Point Prompter} & \textit{Text Prompter} & MN & ONLY & BG & PB \\
            \midrule
                   &        & 59.8 & 32.2 & 32.0 & 37.3 \\
                   & \cmark & 90.4 & 85.0 & 86.5 & 77.2 \\
            \cmark &        & 90.8 & 85.7 & 88.0 & 79.8 \\
            \cmark & \cmark & \textbf{94.0} & \textbf{93.1} & \textbf{95.2} & \textbf{88.1} \\
            \bottomrule
        \end{tabular}
    \vspace{-2mm}
    \end{table}

    \begin{table}[!t]
        \centering
        \caption{Ablation study on two loss terms showing both help improve performance.}
        \label{tab:ab_loss}
        \vspace{-2mm}
        \begin{tabular}{cccccccc}
            \toprule
            $\mathcal{L}_{proto}$ & $\mathcal{L}_{reg}$ & $\mathcal{L}_{con}$ & MN & ONLY & BG & PB \\
            \midrule
                    &         & \cmark  & 93.3 & 91.7 &  93.4   &  86.5  \\
                    & \cmark  & \cmark  & 93.5 & 92.3 &  93.6   &  86.8  \\
            \cmark  &         & \cmark  & 93.8 & 93.1 &  94.5   &  87.8  \\
            \cmark  & \cmark  & \cmark  & \textbf{94.0} & \textbf{93.5} &  \textbf{95.2}   &  \textbf{88.1}  \\
            \bottomrule
        \end{tabular}
    \vspace{-2mm}
    \end{table}

    \begin{table}[!t]
        \centering
        \caption{Contribution of Consistency Loss to Cross-Dataset Generalization. It is essential for improving target performance while maintaining comparable source accuracy.}
        \label{tab:ab_cross}
        \vspace{-2mm}
        \begin{tabular}{lccccccccc}
            \toprule
            \multirow{2}{*}{Method} & Source & \multicolumn{5}{c}{Target}       \\
            \cmidrule(lr){2-2} \cmidrule(lr){3-7} 
             & LVIS   & MN & ONLY & BG & PB & Avg. \\
            \midrule
            P$^3$T   &  39.3   & \textbf{62.4} & \textbf{52.8} &  \textbf{48.7}  & \textbf{37.0} & \textbf{50.2} \\
            $\ -\mathcal{L}_{con}$ &  \textbf{41.1}  & 61.6 & 45.1 & 44.8  & 28.3 & 45.0 \\
            \midrule
            $\Delta$ & \textcolor{blue}{+1.8} & \textcolor{red}{-0.8} & \textcolor{red}{-7.7} & \textcolor{red}{-3.9} & \textcolor{red}{-8.7} & \textcolor{red}{-5.2} \\
            \bottomrule
        \end{tabular}
    \vspace{-3mm}
    \end{table}

    \subsection{Ablation Study}
        \subsubsection{Prompt Patches Selection}
            Fig.~\ref{fig:ratio} plots classification performance by prompt ratio $\alpha$ and target patch selection method. The suffix ``-C'' and ``-R'' indicate critical and random patch selection, respectively. Critical patches are the top $\alpha$ fraction of patches with the highest importance scores, as opposed to our strategy. Random selection uniformly produces inferior performance, while critical selection is optimal at 0.1. However, our method significantly outperforms both from 0.3 onwards, achieving the best performance at 0.5.
            By retaining influential patches and prompting less informative ones, our method enriches input point cloud with a more diverse and informative feature set.

        \subsubsection{Point Prompter and Text Prompter}
            Tab.~\ref{tab:ab_prompters} shows the ablation study on the proposed \textit{Point Prompter} and \textit{Text Prompter} for the object classification task.
            The first row corresponds to the frozen pre-trained model without any \textit{Prompter}, i.e., the zero-shot baseline. 
            The second and third rows evaluate each \textit{Prompter} individually, both yielding notable performance gains, with the \textit{Point Prompter} performing slightly better.
            Finally, combining both \textit{Prompters} represents our method, achieving the best overall performance.

        \subsubsection{Loss Analysis}
            The ablation results of the classification for the two loss terms are shown in Tab.~\ref{tab:ab_loss}. 
            Among them, $\mathcal{L}_{proto}$ is the most critical factor, contributing most significantly to performance improvement, while $\mathcal{L}_{reg}$ also provides modest gains.
            Tab.~\ref{tab:ab_cross} highlights the importance of $\mathcal{L}_{con}$ for cross-dataset generalization. Removing this term leads to a slight increase on source dataset, but results in a substantial drop on all target datasets, indicating its role in promoting generalizability.
            Since LVIS contains a lot of fine-grained categories ($\sim$1k) with a long-tail distribution, regularizing the expressiveness of learnable prompts may limit performance on the source dataset.

        \subsubsection{Module Architecture}
            Tab.~\ref{tab:ab_arc} reports the number of learnable parameters and classification performance across different architectures of the two modules in the \textit{Point Prompter}: Prompt Encoder $h_e(\cdot)$ and Offset Generator $h_o(\cdot)$. Our default design---3 EdgeConvs for $h_e$ and 1 EdgeConv for $h_o$---achieves the best trade-off between accuracy and efficiency, outperforming both shallower and heavier variants.

    \begin{table}[!t]
        \centering
        \caption{Effect of the number of learnable parameters in each module. The default design offers the best accuracy with high parameter efficiency.}
        \label{tab:ab_arc}
        \vspace{-2mm}
        \begin{tabular}{llcccc}
            \toprule
            Module & Architecture & \#LP & BG & PB\\
            \midrule
            \multirow{3}{*}{Prompt Encoder$\ h_e(\cdot)$} & 1 EdgeConv & 1.1 & 94.3  & 87.5 \\
                & 3 EdgeConvs    & 2.0  & \textbf{95.2}  & \textbf{88.1} \\
                & 1 Transformer  & 2.5  & 94.3  & 87.1 \\
            \midrule
            \multirow{3}{*}{Offset Generator$\ h_o(\cdot)$}  & 1 MLP & 1.7 & 94.3  & 87.2 \\
              & 1 EdgeConv     & 2.0  & \textbf{95.2}  & \textbf{88.1} \\
              & 3 EdgeConvs    & 2.9  & 94.8  & 87.5 \\
           \bottomrule
        \end{tabular}
    \vspace{-2mm}
    \end{table}

    \begin{figure}[!t]
        \centering
        \includegraphics[width=1.\linewidth]{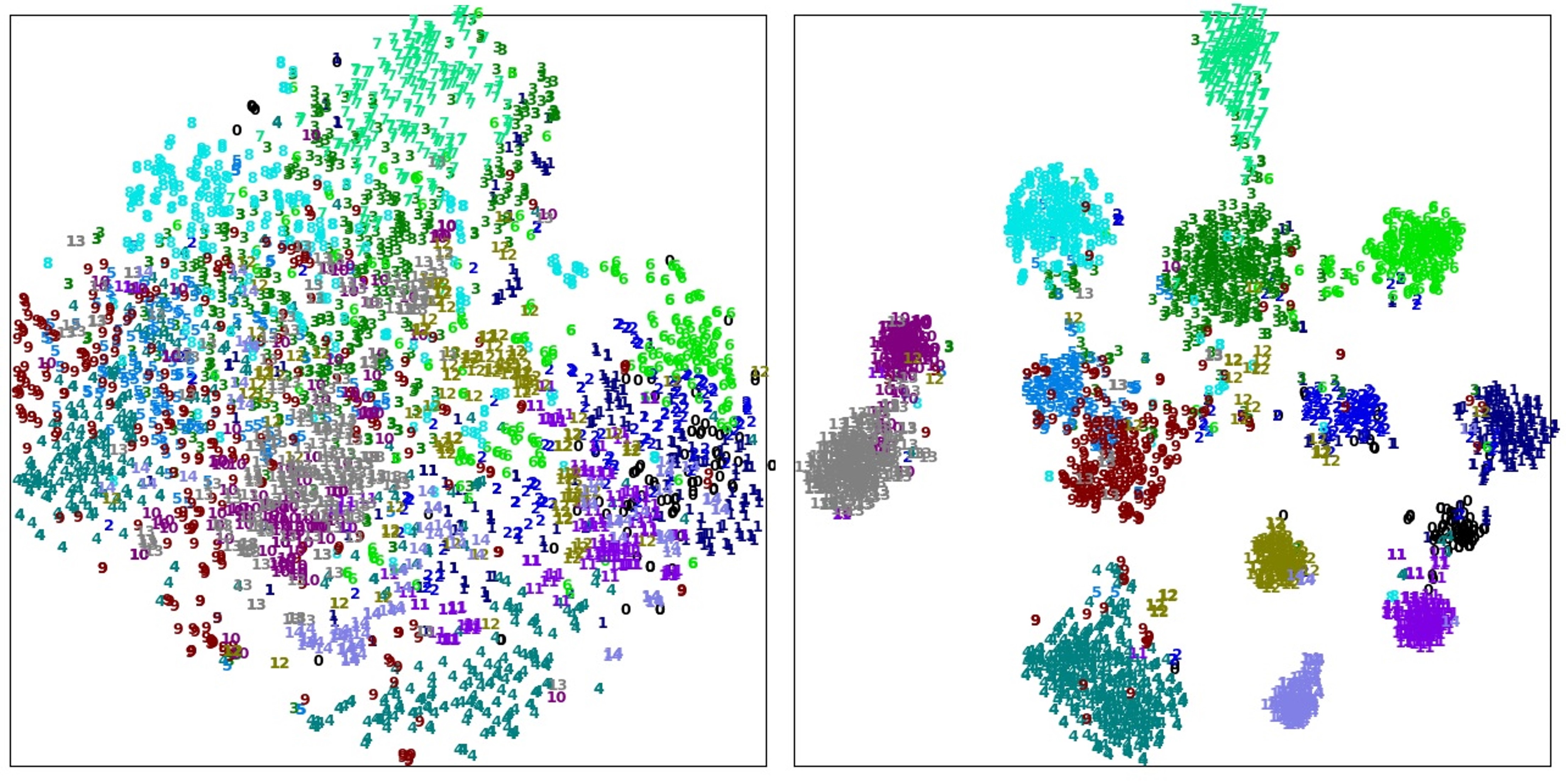}
        \vspace{-7mm}
        \caption{The t-SNE visualization of the 3D embeddings from PB dataset before (\textbf{left}) and after (\textbf{right}) fine-tuning. Embeddings with the same color correspond to the same category.}
        \label{fig:tsne}
    \vspace{-3mm}
    \end{figure}

    \subsection{Qualitative Result}
        Fig.~\ref{fig:tsne} illustrates the t-SNE visualization of 3D embeddings from PB before and after fine-tuning. Before fine-tuning, category distinctions were unclear and intra-category variance was high. After fine-tuning, embeddings from the same category form well-defined clusters.
        This indicates that our method encourages more discriminative feature representations, which contribute to improved performance.

\section{CONCLUSION}
    We propose Prototypical Point-level Prompt Tuning (P$^3$T), an efficient and effective prompt tuning method for 3D vision-language models (VLMs). P$^3$T applies point-level prompting directly to the input point cloud and employs consistency regularization on learnable prompts in the text branch to integrate general textual knowledge.
    By keeping the pre-trained models entirely frozen, it enables adaptation to downstream tasks without compromising generalizability.
    Furthermore, the prototypical loss reduces intra-category variance in the embedding space, resulting in better alignment---particularly beneficial for noisy real-scanned datasets.
    Extensive experiments demonstrate that P$^3$T not only achieves state-of-the-art performance in 3D recognition tasks, but also significantly improves generalization in cross-dataset setting.
    Overall, P$^3$T offers a scalable and robust solution for efficiently deploying large-scale 3D VLMs in diverse real-world applications.


\section*{ACKNOWLEDGMENT}
    This work was supported by the 2024 Research Fund of the University of Seoul for Jiyoung Jung. Also, this research was supported by the National Research Foundation of Korea(NRF) grant funded by the korea government(MSIT) for Geunyoung Jung(NO. RS-2022-NR068754, RS-2025-24523036).

\bibliography{IEEEabrv}
\bibliographystyle{IEEEtran}

\end{document}